\documentclass[10pt,journal,compsoc]{IEEEtran}
\ifCLASSOPTIONcompsoc
  \usepackage[nocompress]{cite}
\else
  \usepackage{cite}
\fi

\usepackage{graphicx}
\usepackage{amsmath}
\usepackage{url}
\usepackage{calc}
\usepackage{subfig}
\usepackage{amssymb}
\usepackage{amstext}
\usepackage{amsmath}
\usepackage{graphicx}
\usepackage{amsfonts}       
\usepackage{nicefrac}       
\usepackage{multicol}
\usepackage{pslatex}
\usepackage{apalike}
\usepackage{ragged2e}
\usepackage{mathtools}
\usepackage{multirow}
\usepackage{multicol}
\usepackage{tabularx}
\usepackage{float}
\usepackage{pslatex}
\usepackage[table]{xcolor}
\graphicspath{{images/VISAPP_old}, {images/al_report}}
\usepackage{hyperref}
\hypersetup{
           breaklinks=true,   
           colorlinks=true,   
           pdfusetitle=true,  
           linkcolor=blue,
           citecolor=blue,
           urlcolor=blue
        }

\begin{document}
%
\title{
Evaluating the effect of data augmentation and BALD heuristics on distillation of Semantic-KITTI dataset
}

%
%
%
%

\author{Ngoc Phuong Anh Duong,
        Alexandre Almin, Léo Lemarié
        and~B Ravi Kiran
\IEEEcompsocitemizethanks{\IEEEcompsocthanksitem 
Machine Learning, Navya
.\protect\\
E-mail: see first.lastname@nsvya.tech}
\thanks{Preprint October 2022.}}

\maketitle

\begin{abstract}
Active Learning (AL) has remained relatively unexplored for LiDAR perception tasks in autonomous driving datasets. In this study we evaluate Bayesian active learning methods applied to the task of dataset distillation or core subset selection (subset with near equivalent performance as full dataset). We also study the effect of application of data augmentation (DA) within Bayesian AL based dataset distillation. We perform these experiments on the full Semantic-KITTI dataset. We extend our study over our existing work \cite{visapp22Duong} only on 1/4th of the same dataset. Addition of DA and BALD have a negative impact over the labeling efficiency and thus the capacity to distill datasets. We demonstrate key issues in designing a functional AL framework and finally conclude with a review of challenges in real world active learning.
\end{abstract}

\begin{IEEEkeywords}
Active Learning \and Semantic Segmentation \and Dataset distillation \and Data augmentation
\end{IEEEkeywords}



%

\section{Introduction} 
Autonomous driving perception datasets in the point cloud domain including
Semantic-KITTI \cite{behley2019semantickitti} and nuScenes \cite{caesar2020nuscenes} 
provide a large variety of driving scenarios \& lighting conditions, along with 
variation in the poses of on-road obstacles. Large scale datasets are created across
different sensor sets, vehicles and sites. There are multiple challenges in creating a
functional industrial grade autonomous driving dataset \cite{uricar2019challenges}. 
During the phase of creating a dataset, these following key steps are generally
followed:
\begin{itemize}
    \item Defining the operation design domain of operation of the perception models. This involves parameters (but are not limited to) such as minimum \& maximum distance, speed \& illumination, classes to be recognized in the scene, relative configurations of objects or classes w.r.t ego-vehicle,  
    \item Fixing sensors (camera, LiDAR, RADAR, ...) suite, with their number \& parameters,
    \item Choosing target vehicles and locations over which logs should be collected,
    \item Collecting logs given the ODD parameters, vehicles and sites,
    \item Selecting key frames and samples to annotate,
    \item Creating a labeled dataset by using human experts,
    \item Incrementally improving the dataset for a given model by using active learning strategies.
\end{itemize}

These large-scale point clouds datasets have high redundancy due to temporal and spatial correlation. Redundancy makes training Deep Neural Network (DNN) architectures costlier for very little gain in performance. This redundancy is mainly due to the temporal correlation between point clouds scans, the similar urban environments and the symmetries in the driving environment (driving in opposite directions at the same location). Hence, data redundancy can be seen as the similarity between any pair of point clouds resulting from geometric transformations as a consequence of ego-vehicle movement along with changes in the environment. Data augmentations (DA) are transformations on the input samples that enable DNNs to learn invariances and/or equivariances to said transformations \cite{anselmi2016invariance}. DA  provides a natural way to model the geometric transformations to point clouds in large-scale datasets due to ego-motion of the vehicle.

Active Learning (AL) is an established field that aims at interactively annotating unlabeled samples guided by a human expert in the loop. With existing large datasets, AL methods could be used to find a core-subset with equivalent performance w.r.t a full dataset. This involves iteratively selecting subsets of the dataset that greedily maximises model performance. As a consequence, AL helps reduce annotation costs, while preserving high accuracy. AL distills an existing dataset to a smaller subset, thus enabling faster training times in production. It uses uncertainty scores obtained from predictions of a model or an ensemble to select informative new samples to be annotated by a human oracle. Uncertainty-based sampling is a well-established component of AL frameworks today \cite{settles2009active}. 

This extended study following our paper \cite{visapp22Duong} demonstrates results on full dataset distillation on the Semantic-KITTI dataset. In the original study we evaluated the effect of data augmentation on the quality of heuristic function to select informative samples in the active learning loop. We demonstrated for a 6000 samples train subset and 2000 samples test subset of Semantic-KITTI, the increase in performance of label efficiency that data augmentation methods achieved. 

In the current study we have extended our previous work and performed the dataset distillation over the complete Semantic-KITTI dataset. Contributions include:
\begin{enumerate}
    \itemsep0pt
    \item An evaluation of Bayesian AL methods on the complete Semantic-KITTI dataset using data augmentation as well as heuristics from BAAL libraries \cite{atighehchian2019baal}\cite{atighehchian2020bayesian}. 
    \item We compare the AL study from \cite{visapp22Duong} that is performed on a smaller subset (1/4th) of same dataset with the full study, and demonstrate that the data augmentation schemes as well as BALD Bayesian heuristic have negligible gains over a random sampler. We point out the key issues underlying this poor performance.
    \item A qualitative analysis of how labelling efficiency changes when increasing dataset size.
    \item We also perform an ablation study comparing 2 different models: SqueezeSegV2 and SalsaNext within an AL framework.
    \item A summary of key real-world challenges in active learning over large point cloud datasets.
\end{enumerate}
Like many previous studies on AL, we do not explicitly quantify the amount of redundancy in the datasets and purely determine the trade-off of model performance with smaller subsets w.r.t the original dataset.

\subsection{Related work}

The reader can find details on the major approaches to AL in the following articles: uncertainty-based approaches \cite{gal2017deep}, diversity-based approaches \cite{sener2018active}, and a combination of the two \cite{kirsch2019batchbald}\cite{ash2020deep}. Most of these studies were aimed at classification tasks. Adapting diversity-based frameworks usually applied to a classification, such as \cite{sener2018active}, \cite{kirsch2019batchbald}, \cite{ash2020deep}, to the point cloud semantic segmentation task is computationally costly. This is due to the dense output tensor from DNNs with a class probability vector per pixel, while the output for the classification task is a single class probability vector per image.
Various authors in \cite{kendall2017uncertainties}\cite{golestaneh2020importance}, Camvid \cite{Brostow2009SemanticOC} and Cityscapes\cite{cordts2016cityscapes} propose uncertainty-based methods for image and video segmentation. 
However, very few AL studies are conducted for point cloud semantic segmentation. Authors \cite{wu2021redal} evaluate uncertainty and diversity-based approaches for point cloud semantic segmentation. This study is the closest to our current work.

Authors \cite{birodkar2019semantic} demonstrate the existence of redundancy in CIFAR-10 and ImageNet datasets, using agglomerative clustering in a semantic space to find redundant groups of samples. As shown by \cite{chitta2019training}, techniques like ensemble active learning can reduce data redundancy significantly on image classification tasks. Authors \cite{beck2021effective} show that diversity-based methods are more robust compared to standalone uncertainty methods against highly redundant data. Though authors suggest that with the use of DA, there is no significant advantage of diversity over uncertainty sampling. Nevertheless, the uncertainty was not quantified in the original studied datasets, but were artificially added through sample duplication. This does not represent real word correlation between sample images or point clouds. Authors \cite{hong2020deep} uses DA techniques while adding the consistency loss within a semi-supervised learning setup for image classification task. 

\section{Active learning: Method}

\subsection{Dataset, point cloud representation \& DNN-Model}
Although there are many datasets for image semantic segmentation, 
few are dedicated to point clouds.
The Semantic-KITTI dataset \& benchmark by authors \cite{behley2019semantickitti} provides more than 43000 point clouds of 22 annotated sequences, acquired with a Velodyne HDL-64 LiDAR. Semantic-KITTI is by far the most extensive dataset with sequential information. All available annotated point clouds, from sequences 00 to 10, for a total of 23201 point clouds, are later randomly sampled, and used for our experiments. 

\begin{figure*}[ht]
    \centering
    \includegraphics[width=0.7\textwidth]{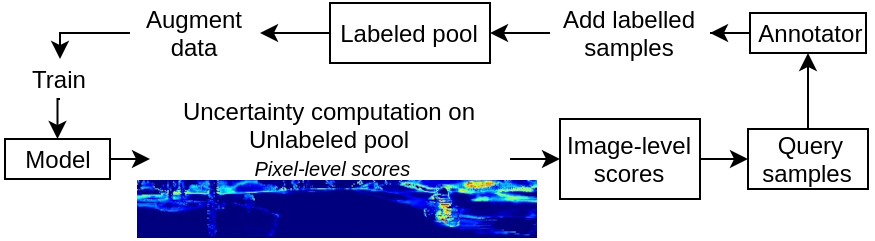}
    \caption{Global flow of active learning on range images from point clouds using uncertainty methods.}
    \label{fig:al_range_image}
\end{figure*}

Among different deep learning models available, we choose SqueezeSegV2 \cite{wu2018squeezesegv2} and SalsaNext\cite{https://doi.org/10.48550/arxiv.2003.03653}, spherical-projection-based semantic segmentation models. While SqueezeSegV2 \cite{wu2018squeezesegv2}  performs well with a fast inference speed compared to other architectures, thus reduces training and uncertainty computation time, SalsaNext is more dense and computationally expensive but shown to have better performance. We apply spherical projection \cite{wu2018squeezesegv2} on point clouds to obtain a 2D range image as an input for the network shown in figure \ref{fig:al_range_image}. To simulate Monte Carlo (MC) sampling for uncertainty estimation \cite{gal2016dropout}, a 2D Dropout layer is added right before the last convolutional layer of SqueezeSegV2. \cite{wu2018squeezesegv2} with a probability of 0.2 and turned only at test time. 

Rangenet++ architectures by authors \cite{milioto2019rangenet++} use range image based spherical coordinate representations of point clouds to enable the use of 2D-convolution kernels. The relationship between range image and LiDAR coordinates is the following:
\begin{equation*}
        \begin{pmatrix} 
        u \\
        v \end{pmatrix} =  
            \begin{pmatrix} 
                \frac{1}{2} [1-\arctan(y,x)\pi^{-1}]\times w \\
                [1-(\arcsin(z\times r^{-1})+{f}_{up})\times {f}^{-1}]\times h 
            \end{pmatrix} ,  
\end{equation*}
 where $(u,v)$ are image coordinates, $(h,w)$ the height and width of the desired range image,  $f =f_{up}+f_{down}$, is the vertical $fov$ of the sensor, and $r = \sqrt{x^2 + y^2 + z^2}$, range measurement of  each  point. The input to the DNNs used in our study are images of size $W \times H \times 4$, with spatial dimensions $W, H$ determined by the FoV and angular resolution, and 4 channels containing the $x, y$ coordinates of points, $r$ range or depth to each point, $i$ intensity or remission value for each point.
 
\subsection{Bayesian AL: Background}

In a supervised learning setup, given a dataset $\mathcal{D}:= \{(\mathbf{x}_1, y_1), (\mathbf{x}_2, y_2), \ldots, (\mathbf{x}_N, y_N) \} \subset \mathcal{X} \times \mathcal{Y}$,
the DNN is seen as a high dimensional function $f_{\omega} : \mathcal{X} \to \mathcal{Y}$ with model parameters $\omega$. A simple classifier maps each input $x$ to outcomes $y$. A good classifier minimizes the empirical risk $l : \mathcal{Y}\times \mathcal{Y} \to
\mathbb{R}$, which is defined with the expectation $R_\text{emp} (f):= \mathbb{P}_{X, Y} [Y \neq f(X)] $. The optimal classifier is one that minimizes the above risk.
Thus, the classifier's loss does not explicitly refer to sample-wise uncertainty but
rather to obtain a function which makes good predictions on average.  We shall use the following terminologies to describe our AL training setup.
\begin{enumerate}
\item \textit{Labeled dataset} $D = \{ \left( \mathbf{x}_i, y_i \right) \}_{i=1}^N$ where $\mathbf{x}_i \in W \times H \times 4$ are range images with 4 input channels, $W, H$ are spatial dimensions, and $y_i \in W \times H \times C$  are one-hot encoded ground truth with $C$ classes. The output of the DNN model is distinguished from the ground truth as $\hat{y}_i$ with the same dimensions. The 4 channels in our case are x, y, z coordinates and LiDAR intensity channel values.
\item \textit{Labeled pool} $L \subset D$ and an unlabeled pool $U \subset D$ considered as a data with/without any ground-truth, where at any AL-step $L \cup U = D$, the subsets are disjoint and restore the full dataset. 
\item Query size $B$, also called a \textit{budget}, to fix the number of unlabeled samples selected for labeling 
\item  Acquisition function, known as heuristic, providing a score for each pixel given the output $\hat{y}_i$ of the DNN model, $f: \mathbb{R}^{W \times H \times C} \to \mathbb{R}^{W \times H}$ 
\item Including the usage of MC iterations where the output of the DNN model could provide several outputs given the same model and input, $\hat{y}_i \in W \times H \times C \times T$ where $T$ refers to the number of MC iterations.
\item \textit{Subset model} $f_L$ is the model trained on labeled subset $L$
\item \textit{Aggregation function} $a: \mathbb{R}^{W \times H \times C \times T} \to \mathbb{R}^+$ is a function that aggregates heuristic scores across all pixels in the input image into a positive scalar value, which is used to rank samples in the unlabeled pool.
\end{enumerate}

Heuristic functions are transformations over the model output probabilities
$p(y|x)$ that define uncertainty-based metrics to rank and select informative
examples from the unlabeled pool at each AL-step. In our previous study \cite{visapp22Duong} we have identified BALD to be a robust heuristic function \cite{houlsby2011bayesian}. It selects samples maximizing information gain between the predictions from model parameters, using MC Iterations.

\subsection{Data augmentations on range images}
\begin{figure*}[ht]
    \centering
    \subfloat[Random dropout mask]{\includegraphics[width=0.47\textwidth]{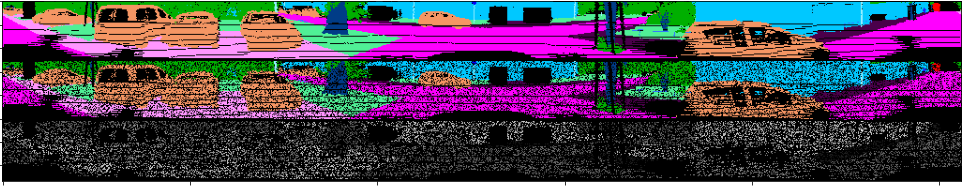}}  \hfill  
    \subfloat[CoarseDropout of Albumentations library]{ \includegraphics[width=0.47\textwidth]{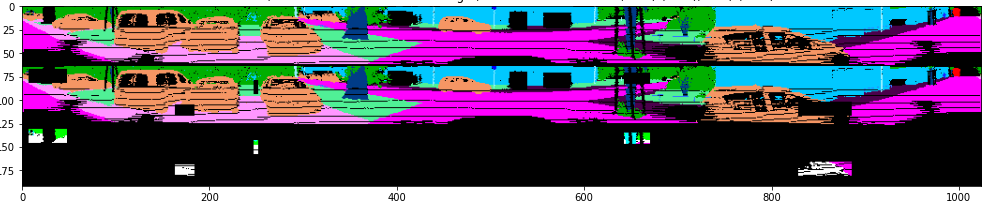}} \\
    \subfloat[Gaussian noise applied on depth channel] { \includegraphics[width=0.47\textwidth]{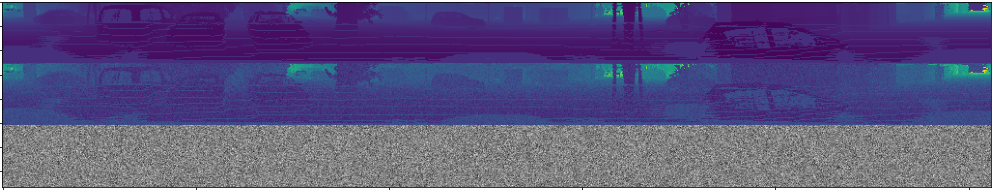}} \hfill  
    \subfloat[Gaussian noise applied on remission channel]{ \includegraphics[width=0.47\textwidth]{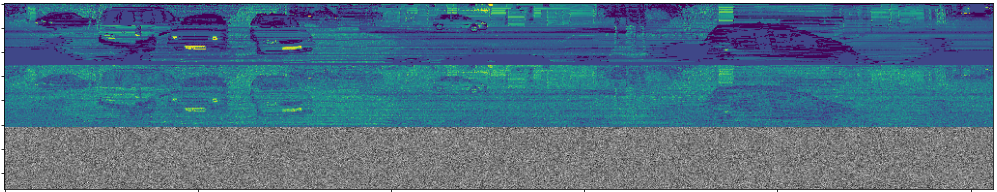}}\\
    \subfloat[Random cyclic shift range image ]{\includegraphics[width=0.47\textwidth]{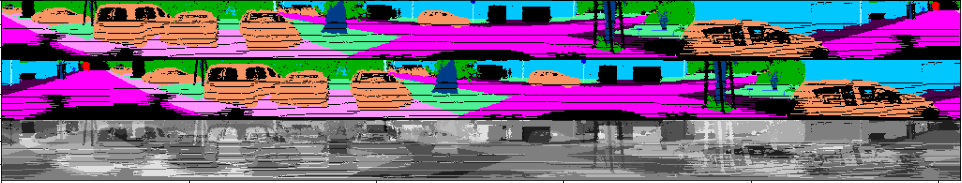}}    \\
    
    \subfloat[Instance Cut Paste]{ \includegraphics[width=0.99\textwidth]{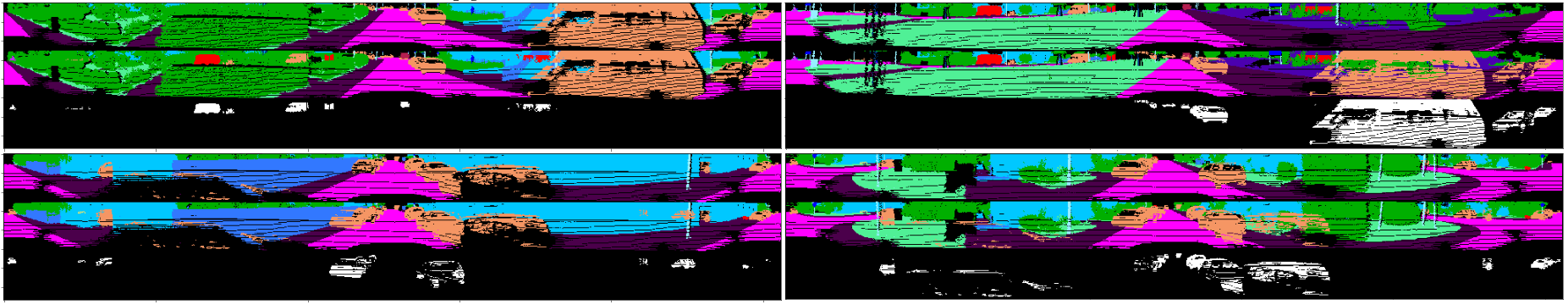}}    
    
    \caption{Before and after applying transformations on Semantic-KITTI. Each image corresponds to a sample such that inner images, from top to bottom, are before and after applying transformations, and the error between them. a, b, c, d are directly used or customized based on Albumentations library \cite{info11020125}}
    \label{fig:da}
       
\end{figure*}

We apply DA directly on the range image projection. We apply the same transformations as in
our previous study \cite{visapp22Duong}, we repeat images of the data augmentation here to visually demonstrate them, in figure \ref{fig:da}.

\subsection{AL Evaluation metrics} 
To evaluate the performance of our experiments we are using the following metrics:
\begin{itemize}
    \item The Mean Intersection over Union (mIoU) \cite{song2016semantic}: $\frac{1}{C}\sum^{C}_{c=1} IoU_c$
    \item \textit{Labeling efficiency}: Authors \cite{beck2021effective} use the labeling
    efficiency (LE) to compare the amount of data needed among different sampling
    techniques with respect to a baseline. In our experiments, instead of
    accuracy, we use mIoU as the performance metric. Given a specific value of
    mIoU, the labeling efficiency is the ratio between the number of labeled
    range images, acquired by the baseline sampling and the other sampling
    techniques.
    \begin{equation}
    \text{LE} = \frac{n_{\text{labeled}\_\text{others}}(\text{mIoU}=a)}{n_{\text{labeled}\_\text{baseline}}(\text{mIoU}=a)}    
    \end{equation}
    The baseline method is usually the random heuristic.
\end{itemize}

\section{Experimental Setup}
In this study we have evaluated the performance of AL based sampling of
a large scale LiDAR dataset Semantic-KITTI. As in \cite{visapp22Duong} we follow a Bayesian
AL loop using MC Dropout. The heuristic computes uncertainty scores for each pixel.
To obtain the final score per range image, we use \emph{sum} as an aggregation
function to combine all pixel-wise scores of an image into a single score. At each
AL step, the unlabeled pool is ranked w.r.t the aggregated score. A new query of
samples limited to the budget size is selected from the ranked unlabeled pool. The
total number of AL steps is indirectly defined by budget size, $n_{AL} = \left| D \right|/B$

In this study we evaluated BALD \cite{houlsby2011bayesian} heuristic, while applying
it with and without DA applied during training time. As mentioned
in Table \ref{tab:setup}, we only use 16000 randomly chosen samples from
Semantic-KITTI over the 23201 samples available. At each training step, we reset
model weights to avoid biases in the predictions, as proven by
\cite{beck2021effective}.

\begin{table*}[]
\centering
\medskip
\resizebox{\textwidth}{!}{
\captionsetup{justification=centering}
\begin{tabular}{|cccccccc|}
\hline
\multicolumn{3}{|c||}{\cellcolor[HTML]{ECF4FF}Data related parameters} & \multicolumn{5}{c|}{\cellcolor[HTML]{ECF4FF}AL Hyper parameters} \\\hline
\textbf{Range image resolution} & \textbf{Total pool size} &
 \multicolumn{1}{c||}{\textbf{Test pool size}}  & \textbf{Init set size} & \textbf{Budget} & \textbf{MC Dropout} & \textbf{AL steps} & \textbf{Aggregation} \\
1024x64 & 6000 &  \multicolumn{1}{c||}{2000} & 240 & 240 & 0.2 & 25 & sum \\\hline
\textbf{Range image resolution} & \textbf{Total pool size} &
 \multicolumn{1}{c||}{\textbf{Test pool size}}  & \textbf{Init set size} & \textbf{Budget} & \textbf{MC Dropout} & \textbf{AL steps} & \textbf{Aggregation} \\
1024x64 & 16241 &  \multicolumn{1}{c||}{6960} & 1041 & 800 & 0.2 & 20 & sum \\\hline
\multicolumn{8}{|c|}{\cellcolor[HTML]{ECF4FF}Hyper parameters for each AL step} \\\hline
\multirow{2}{*}{\textbf{Max train iterations}} & \multirow{2}{*}{\textbf{Learning rate (LR)}} & \multirow{2}{*}{\textbf{LR decay}} & \multirow{2}{*}{\textbf{Weight decay}} & \multirow{2}{*}{\textbf{Batch size}} & \multicolumn{3}{||c|}{\textbf{Early stopping}} \\\cline{6-8}
 &  &  &  &  & \multicolumn{1}{||c}{\textbf{Evaluation period}} & \textbf{Metric} & \textbf{Patience} \\
100000 & 0.01 & 0.99 & 0.0001 & 16 & \multicolumn{1}{||c}{500} & train mIoU & 15\\
\hline
\end{tabular}
}
\caption{Active learning (AL) parameter setup (first line for \cite{visapp22Duong} result).}
\label{tab:setup}
\end{table*}

We evaluate LE mIoU as our metric on the test set of 6960 samples. For quicker training, we use early stopping based on the stability of training mIoU over $\textit{patience}  * \textit{evaluation}\_\textit{period}$ iterations.

\subsection{Analysis of results}
\begin{figure}[ht]
    \centering
    \includegraphics[width=0.75\linewidth]{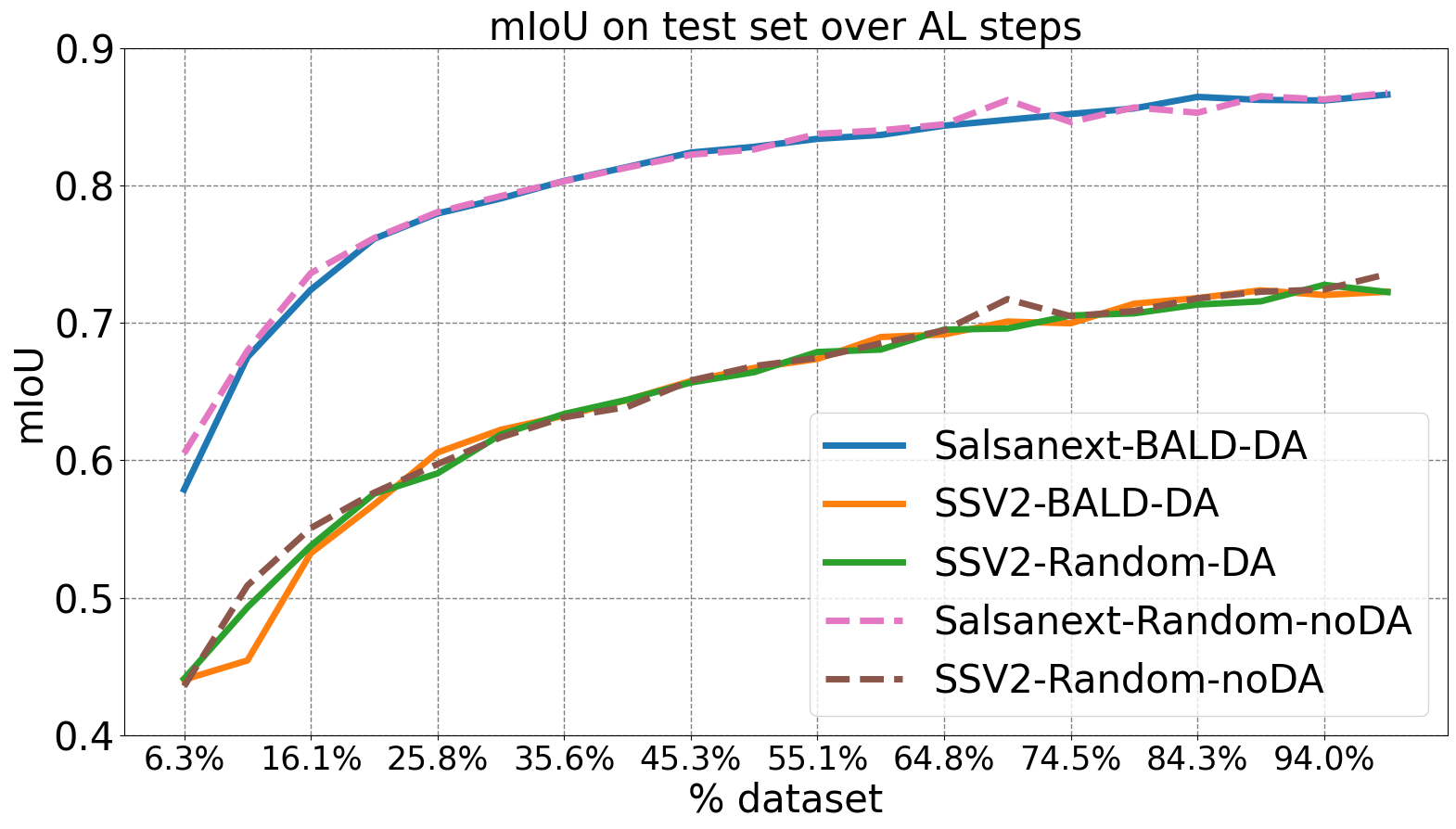}
    \caption{Mean IoU vs dataset size (in percentage) using random and BALD heuristic based samplers under the effect of data augmentation.}
    \label{fig:full_sk_da_results_miou}
\end{figure}

The figure \ref{fig:full_sk_da_results_miou}, demonstrates all the trainings performed on the full Semantic-KITTI dataset. The models SalsaNext (SN) and SqueezeSegV2 (SSV2) were chosen to be evaluated. Both models use rangenet based representations as mentioned earlier. The SSV2 model was trained with and without data augmentation, using a baseline random sampler, the model was also evaluated with BALD heuristic function. SN model was mainly evaluated in the AL framework using data augmentation and with BALD.

The goals here have been to compare the effect of:
\begin{itemize}
    \item Data augmentation on the label efficiency over a large dataset.
    \item Difference in model capacity (SN with 6.7M parameters vs SSV2 with 1M parameters) .
    \item Difference in performance of random sampler vs the BALD heuristic.
\end{itemize}

\begin{figure}
    \centering
    \includegraphics[width=0.475\linewidth]{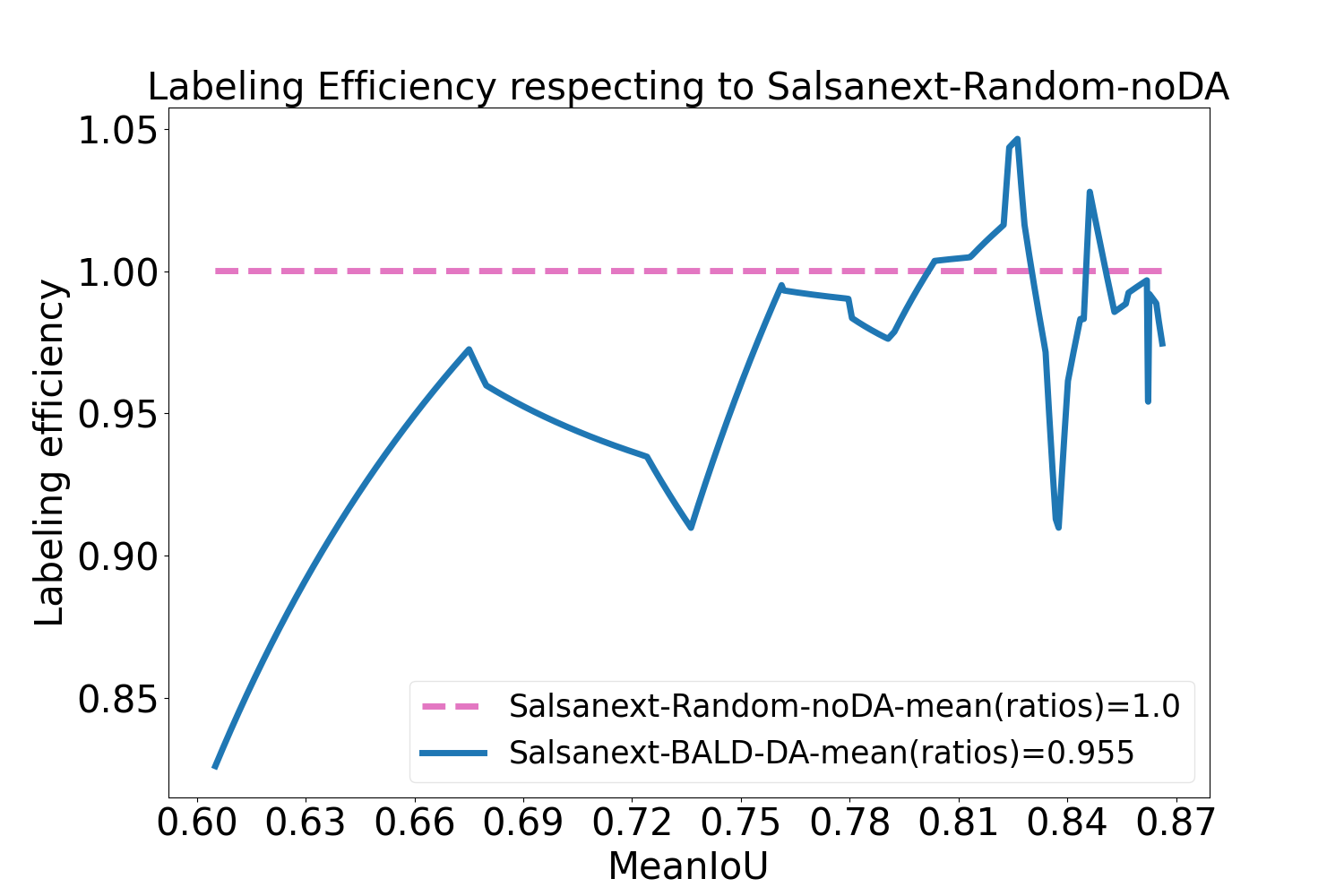}
    \includegraphics[width=0.475\linewidth]{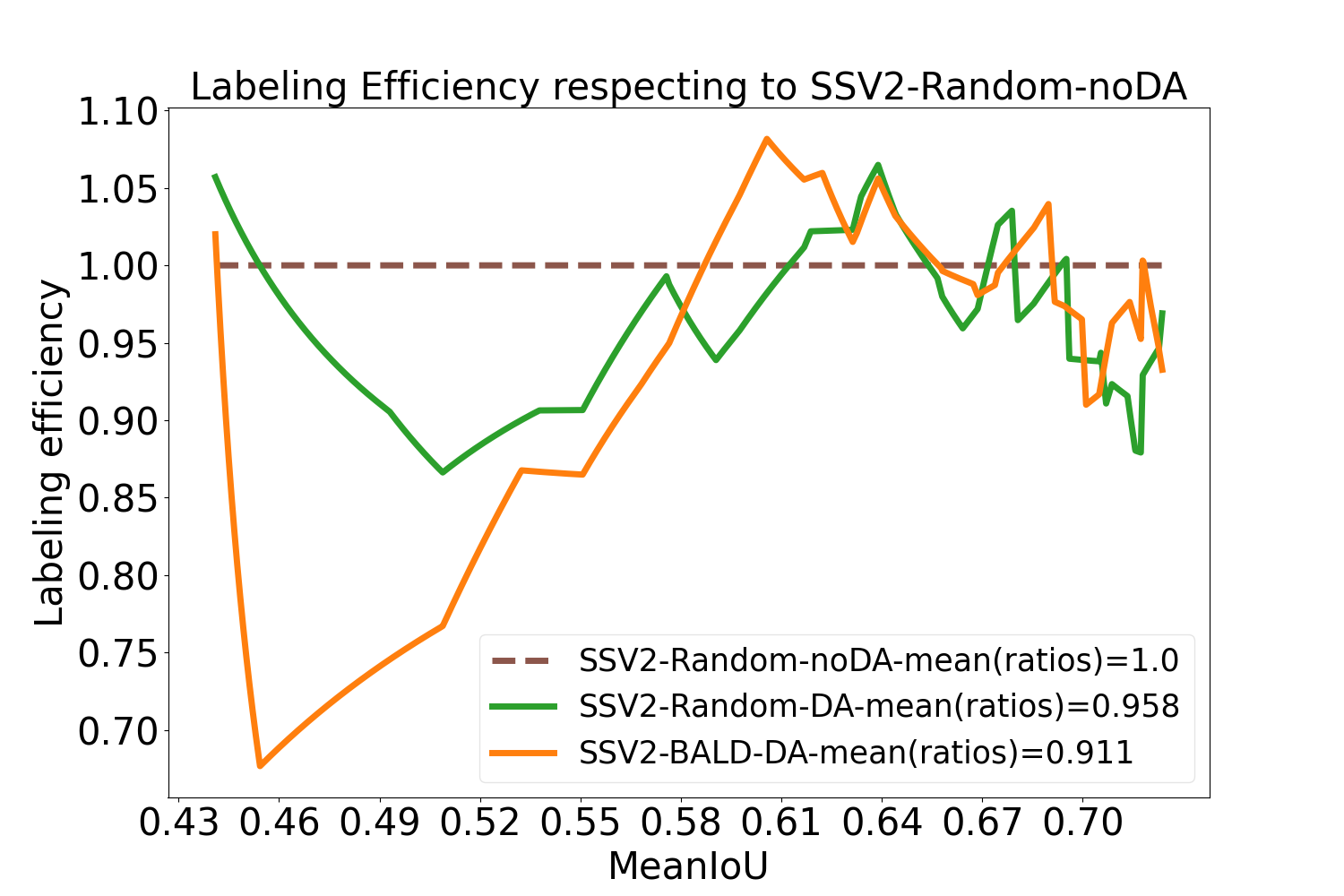}
    \caption{SalsaNext \& SSV2 models evaluated w.r.t the Random sampling baseline. The plots demonstrate the label efficiency.}
    \label{fig:full_sk_SN_SSV2_LE}
\end{figure}

\subsection{Class based learning efficiency}
 In this section we study the variation of ClassIoU for all classes in the Semantic-KITTI dataset. Along with this  we also study the variation from mIoU of the fully supervised (FS) model, expressed by: 
 \begin{equation}
 \label{eq:delta_ciou}
     \Delta cIOU[i] = cIoU[i] - mIoU_{\text{FS}}
 \end{equation}
where $i$ is the index of the active learning loop (or the perception of dataset). This score basically subtracts the mean performance away from the class IoU scores to demonstrate deviations from the mean. The objective of such a measure is to demonstrate the rate at which each class IoU reaches its maximum contribution.

The $\Delta cIOU[i]$ score represents the deviation of the class IoU from the model's full supervised mean performance. When this deviation is positive and large, these represent classes that have been learnt efficiently at the AL-Step i, while negative scores represent classes that have been learnt poorly. This score enables us to determine visually when each class reaches its maximum performance and decide when any incremental addition of samples would have little impact on the final cIoU.

 \begin{figure*}
     \centering
     \includegraphics[width=0.475\linewidth]{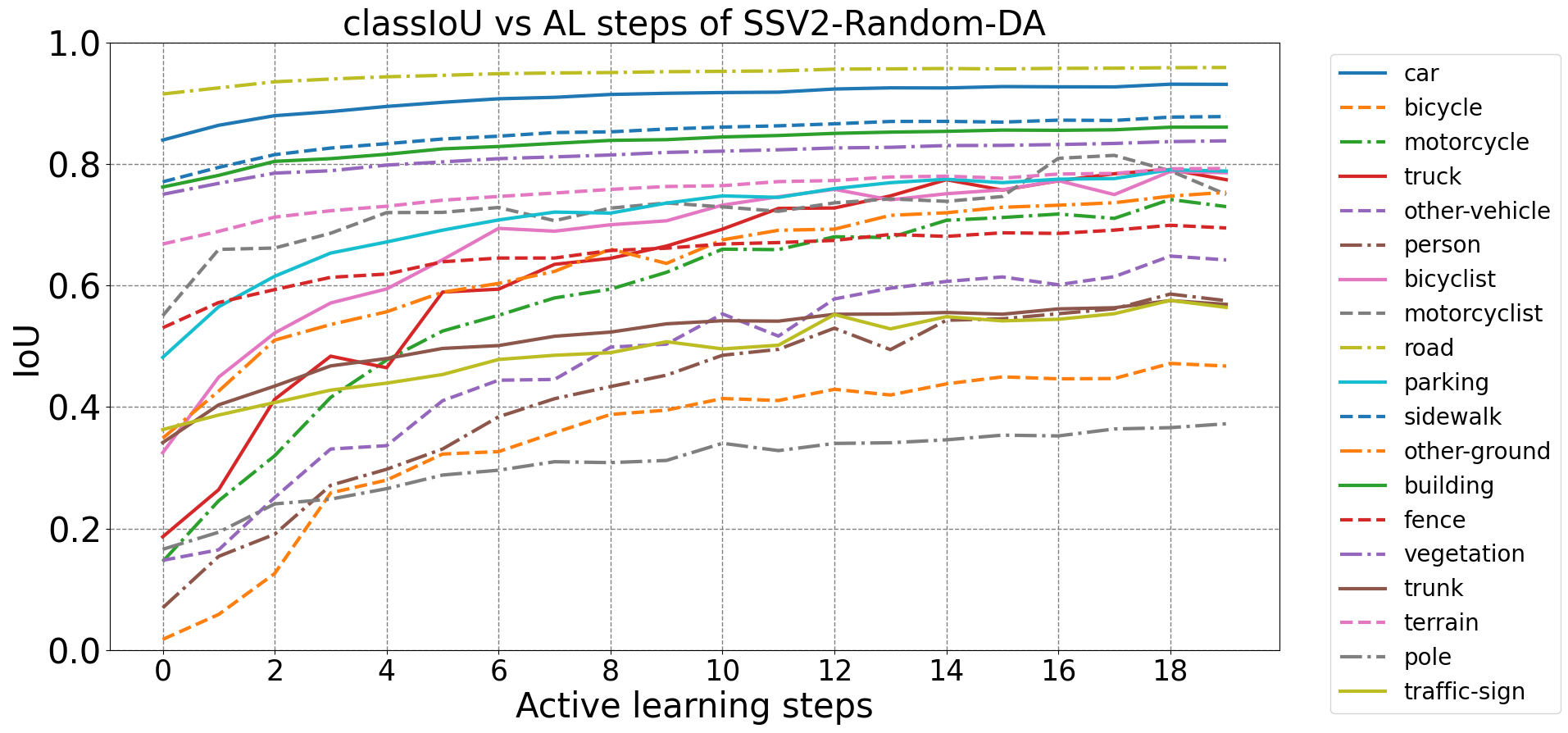}
     \includegraphics[width=0.475\linewidth]{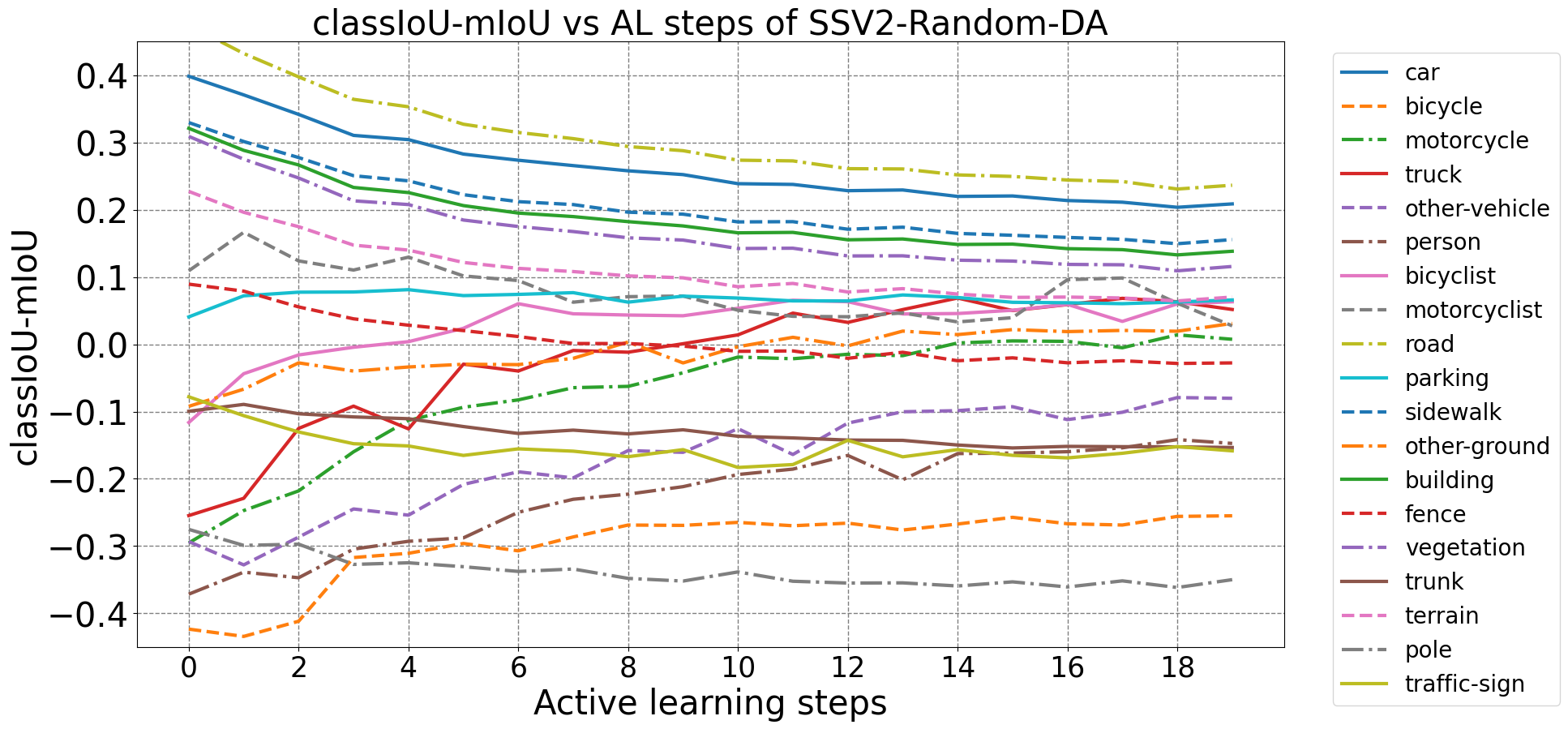}
     \includegraphics[width=0.475\linewidth]{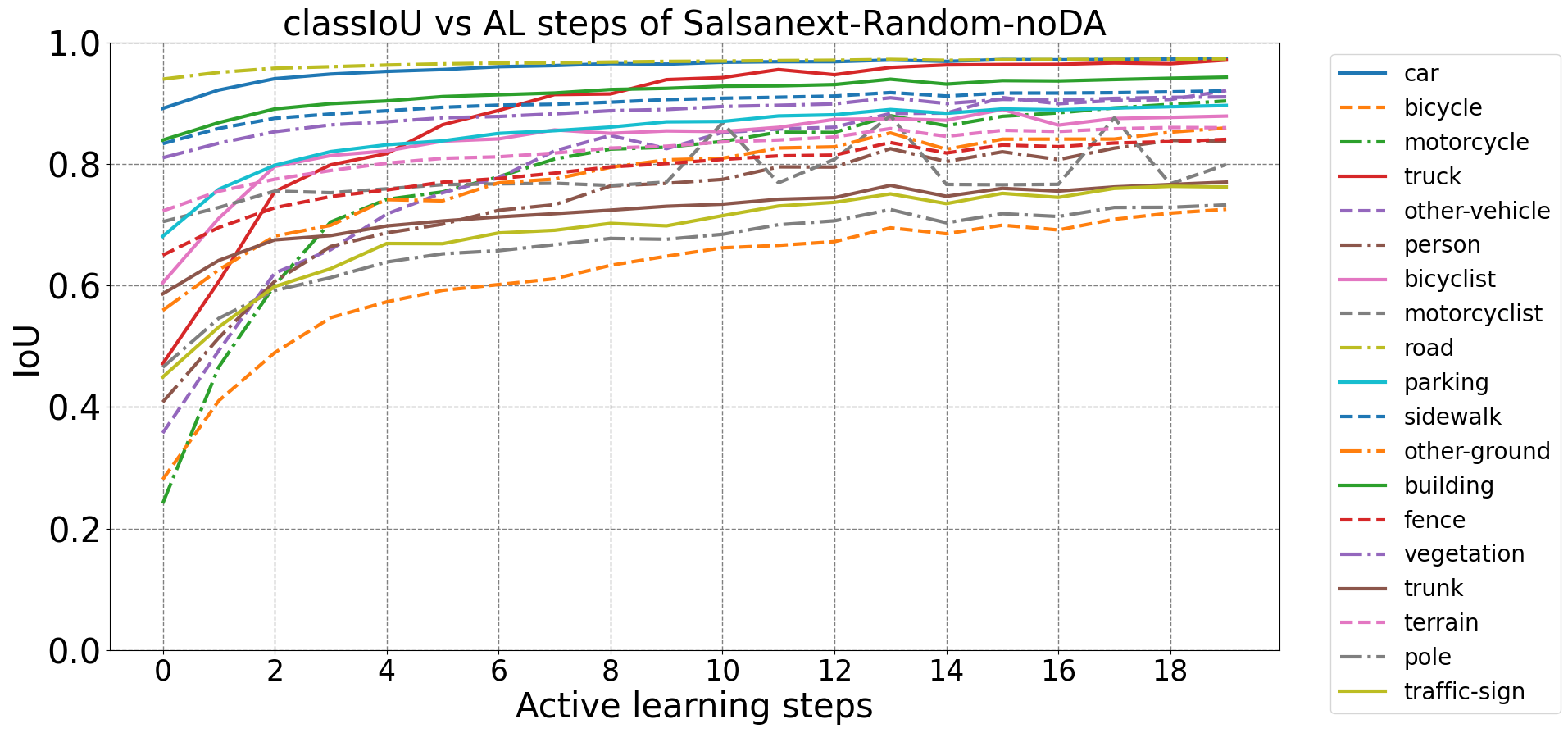}
     \includegraphics[width=0.475\linewidth]{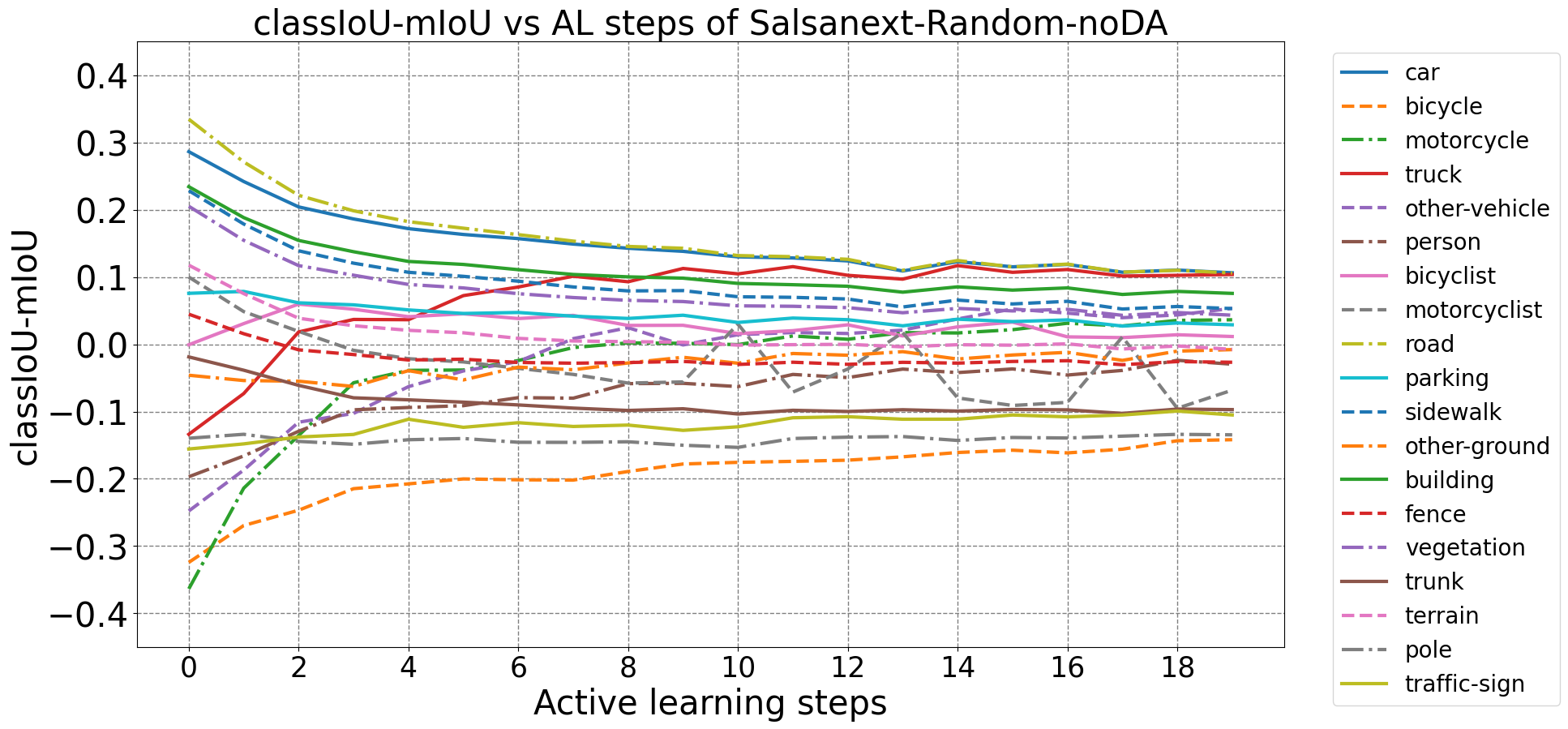}
     \caption{Top Left: Class IoU for SSV2 model Top Right: Deviation from mean IoU from equation \ref{eq:delta_ciou}. Bottom, the same for the SalsaNext model.}
     \label{fig:cIoU_SSV2_SN_da_cIoU}
 \end{figure*}
 
 \textbf{Class-frequency and ClassIoU}: In figure \ref{fig:cIoU_SSV2_SN_da_cIoU} we observe that several majority classes like ROAD, TERRAIN, BUILDING and VEGETATION do not demonstrate any large change in their IoUs after the 6th AL step, while classes such as BICYCLE, POLE, PERSON and OTHER VEHICLE are the slowest to learn their maximum performance. The majority classes produce a positive deviation from mIoU while the least frequent classes produce a negative deviation from mIoU.
 
\begin{figure*}[ht]
     \centering
     \includegraphics[width=0.475\linewidth]{images/full_sk_results/SSV2-Random-DA_diff.png}
     \includegraphics[width=0.475\linewidth]{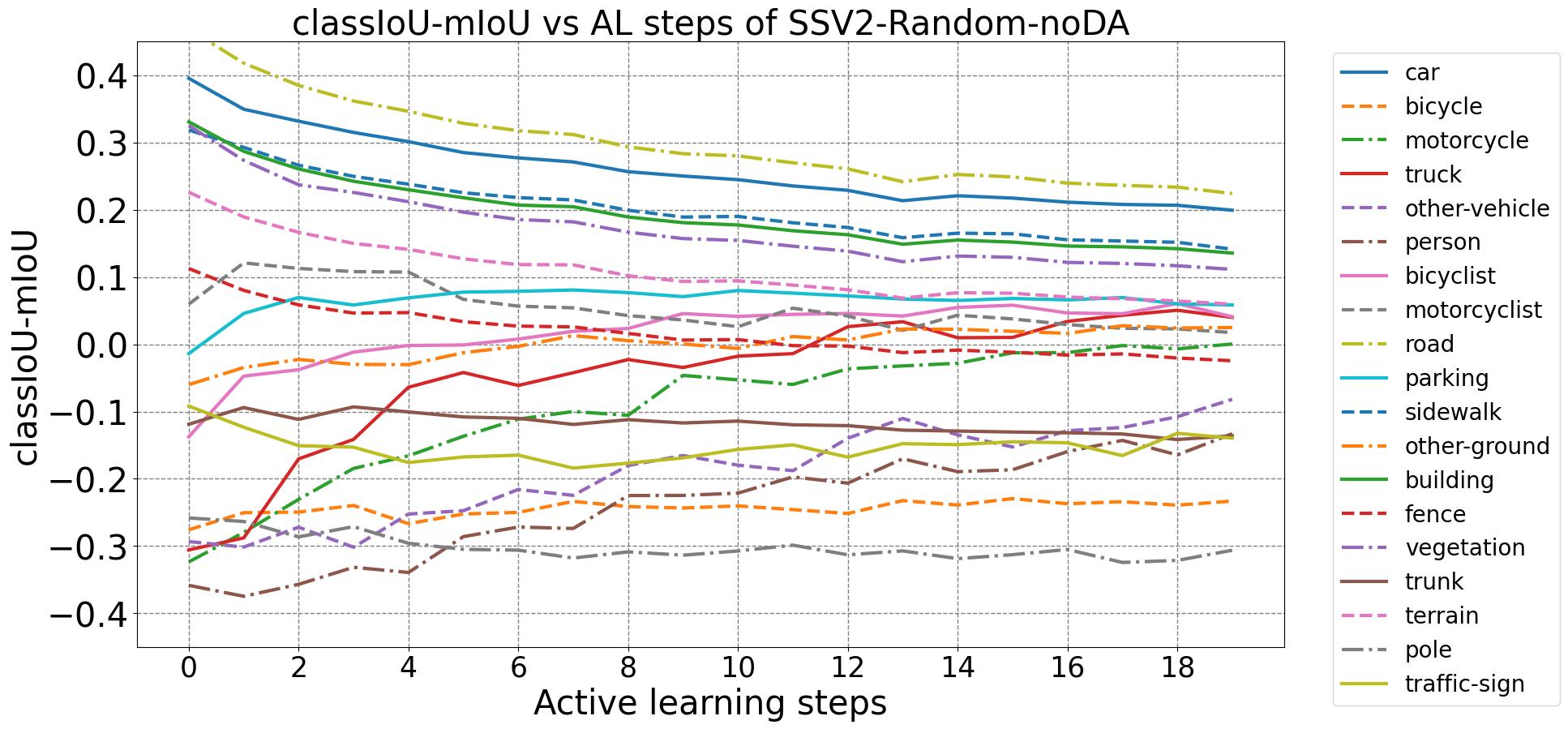}
     \caption{Deviation of Class IoU from mIoU for the SSV2 model: with and without data augmentation applied.}
     \label{fig:cIoU_SSV2_da_noda}
 \end{figure*}

 \textbf{DA effect on ClassIoU}:
 In figure \ref{fig:cIoU_SSV2_da_noda} we plot the $\Delta cIoU[i]$ scores to demonstrate how fast different classes are learnt over the AL loop (different subset sizes of the full dataset). With this plot we would like to observe the change in sample complexity for each class with and without data augmentations using baseline random sampling.

\begin{figure*}[ht]
     \centering
     \includegraphics[width=0.475\linewidth]{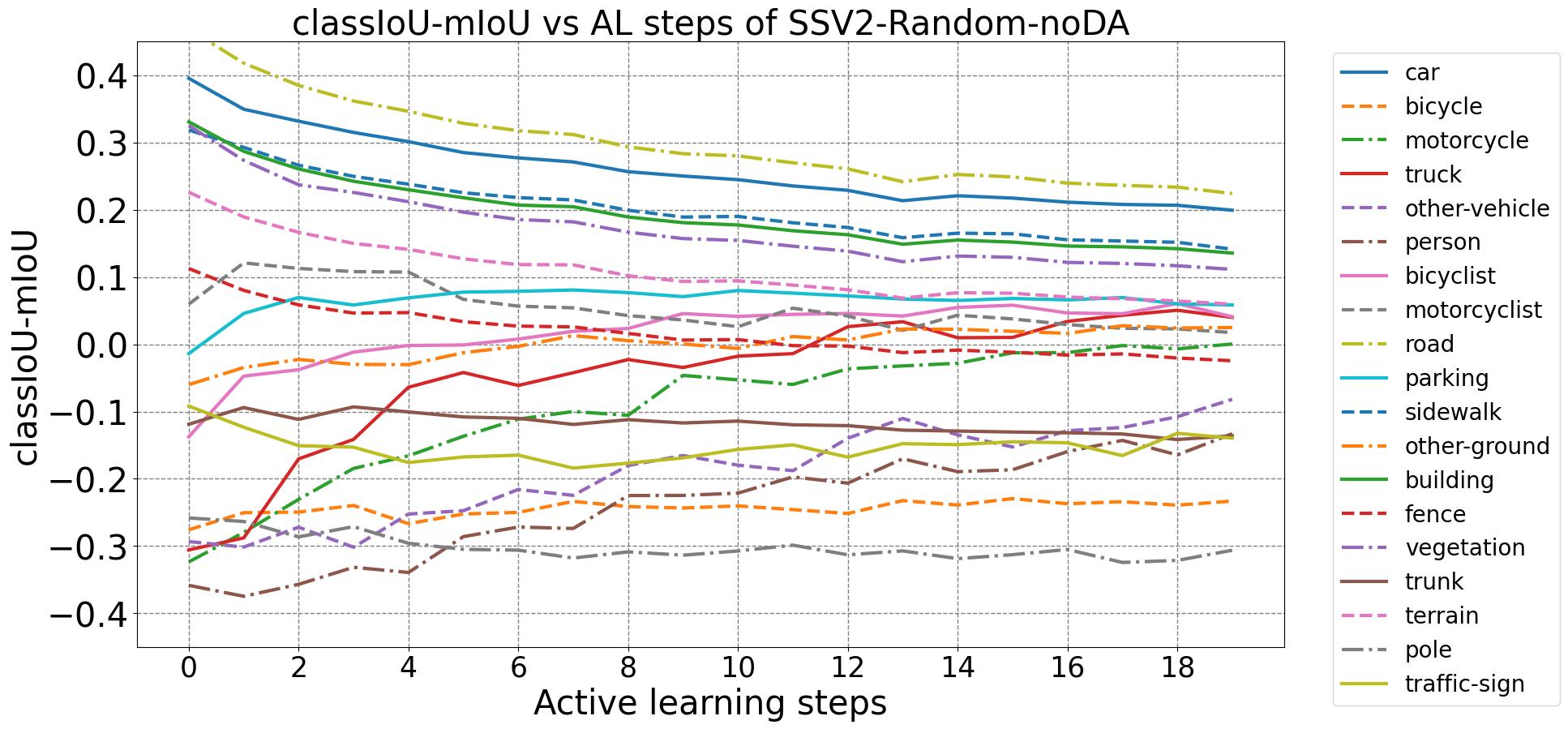}
     \includegraphics[width=0.475\linewidth]{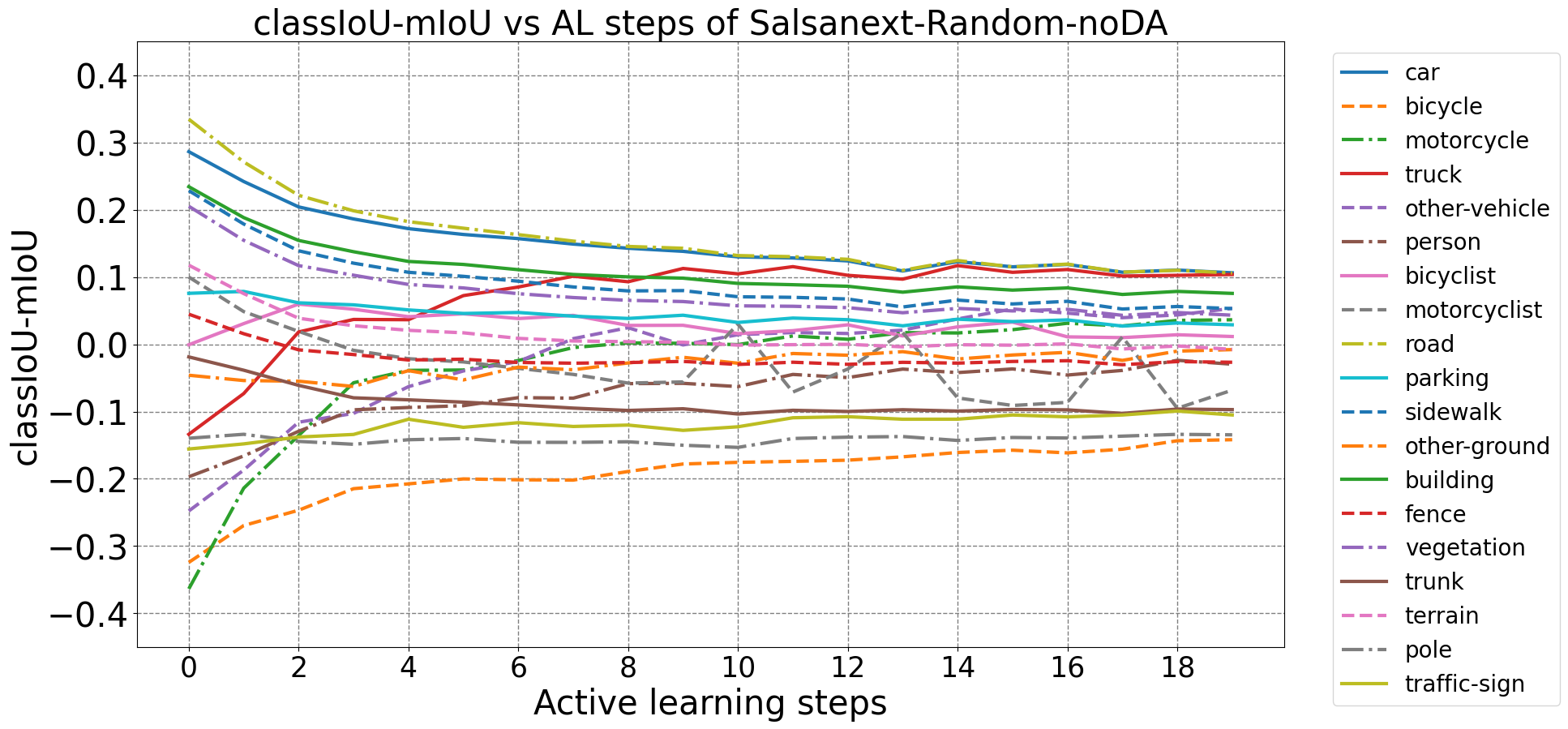}
     \caption{Deviation of ClassIoU from mIoU comparison: Between SSV2 and SN models.}
     \label{fig:cIoU_SSV2_SN_noDA}
 \end{figure*}

\textbf{Model complexity on ClassIoU}:
We evaluate two models on the Semantic-KITTI dataset: SSV2 model vs SalsaNext model. 
We compare the performances of these models in terms of how fast they learn different classes, see figure \ref{fig:cIoU_SSV2_SN_noDA}. We observe that majority of classes have had a large jump within the first 3-4 AL steps for the SalsaNext model, demonstrating how model capacity plays an important role in the sample complexity of learning certain classes.

\subsection{Dataset size growth: 1/4 Semantic-KITTI vs full Semantic-KITTI}

In our previous study on Semantic-KITTI \cite{visapp22Duong} we have evaluated the
performance of AL methods while applying data augmentation on 1/4th subset of the
Semantic-KITTI dataset. As expected, DA sampled harder samples by eliminating
similar samples learnt by invariance. That is to say, models trained with DA are
prone to select samples different from the trained samples and their
transformations, thus reducing redundancy in the selection.

\begin{figure}[ht]
    \centering
    \includegraphics[width=0.45\linewidth]{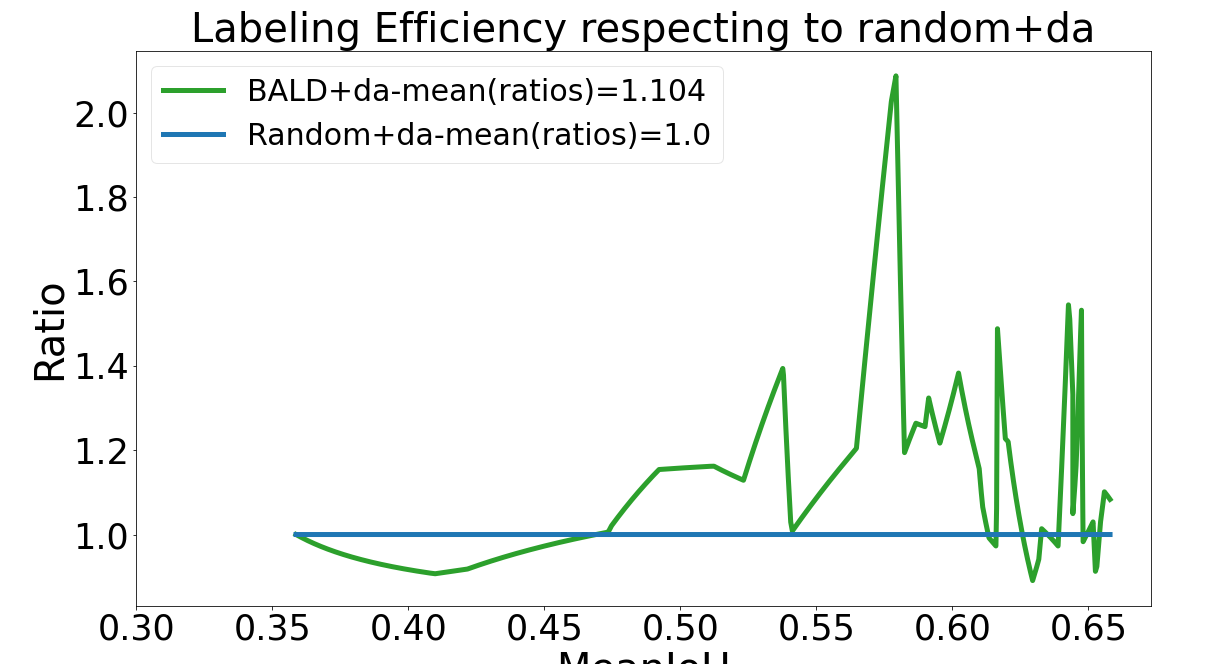}
    \includegraphics[width=0.45\linewidth]{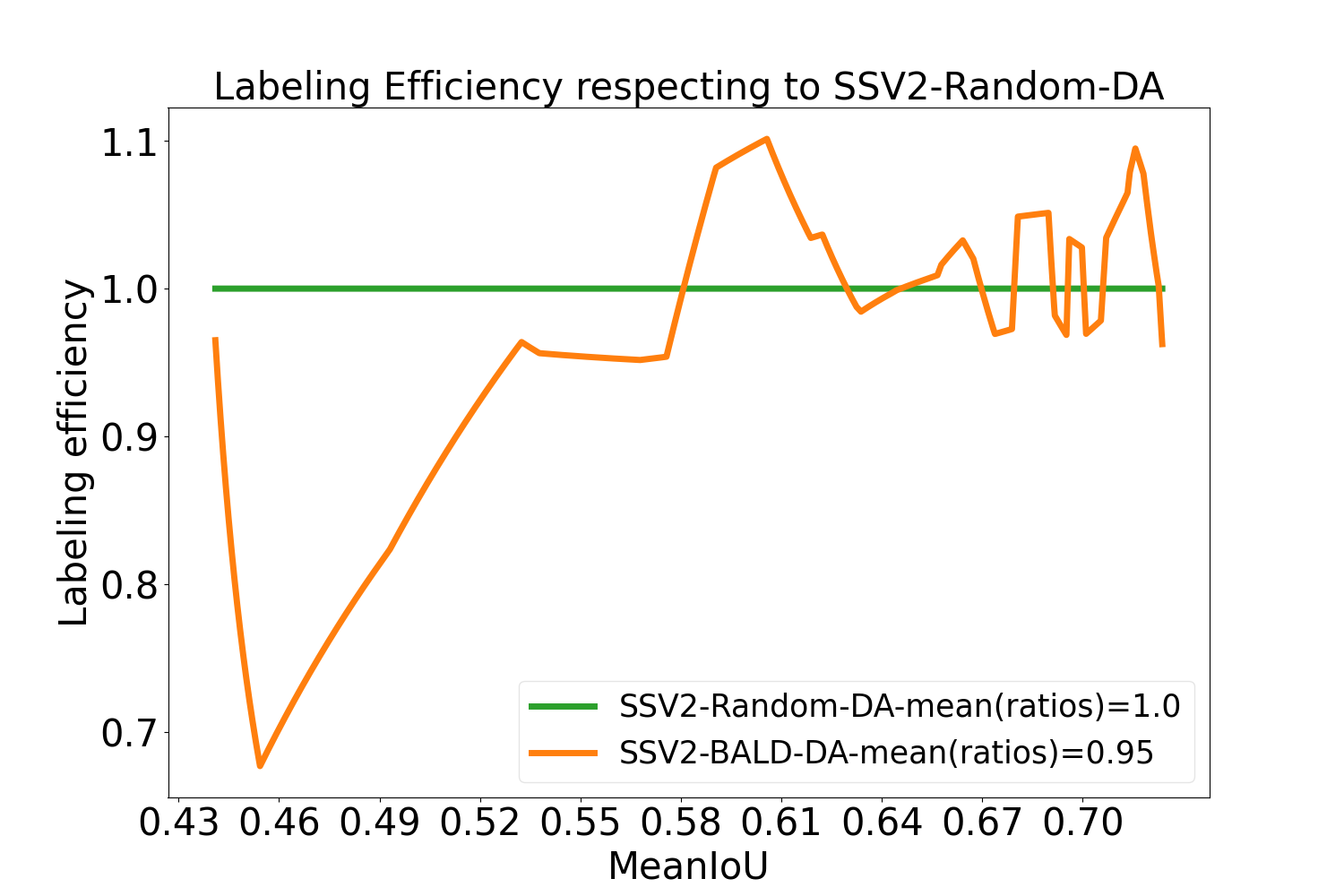}
    \caption{Comparison of the BALD performance. Left: AL training on Semantic-KITTI subset 1/4 from \cite{visapp22Duong}, Right: AL training on Semantic-KITTI full dataset from this study. We observe that the two LE are poor in the initial steps, with subsequent increases in the future AL steps.}
    \label{fig:comparing_datasetsize_LE_BALD}
\end{figure}

On the other hand, although DA significantly reduce redundancy in 1/4 Semantic-KITTI experiments, there is almost no increment in the gain by DA in full Semantic-KITTI experiments. This can be seen in figure \ref{fig:comparing_datasetsize_LE_BALD}
and could be mainly attributed to the following reasons (we hypothesize): 
\begin{enumerate}
    \item Larger dataset includes larger amount of near similar or redundant samples, but all standalone uncertainty-based methods still encounter the problem of high-score similar samples. Semantic-KITTI is a sequential dataset and thus contain similar scans due to temporal correlation. Thus random sampling breaks this redundancy, while methods like BALD are unable to do so due to the above said reasons. Potential solutions for this problem are mentioned in future work and challenges section below.
    \item The gain by DA is getting smaller as dataset size increases because some augmented samples can be found in larger dataset (when dataset is getting larger) or test domain.
    \item From the hypothesis, uncorrelated-to-real-world DA can increase uncertainty in predictions of unnecessary sample points, leading wrong prediction and impractical sampling selection.
\end{enumerate}

\subsection{t-SNE problem analysis}
t-SNE \cite{tsnearticle} is a technique used to visualize high-dimensional data in low-dimensional spaces such as 2D or 3D by minimizing Kullback-Leibler divergence between the low-dimensional distribution and the high-dimensional distribution.

\begin{figure*}[ht]
    \centering
    \includegraphics[width=0.85\textwidth]{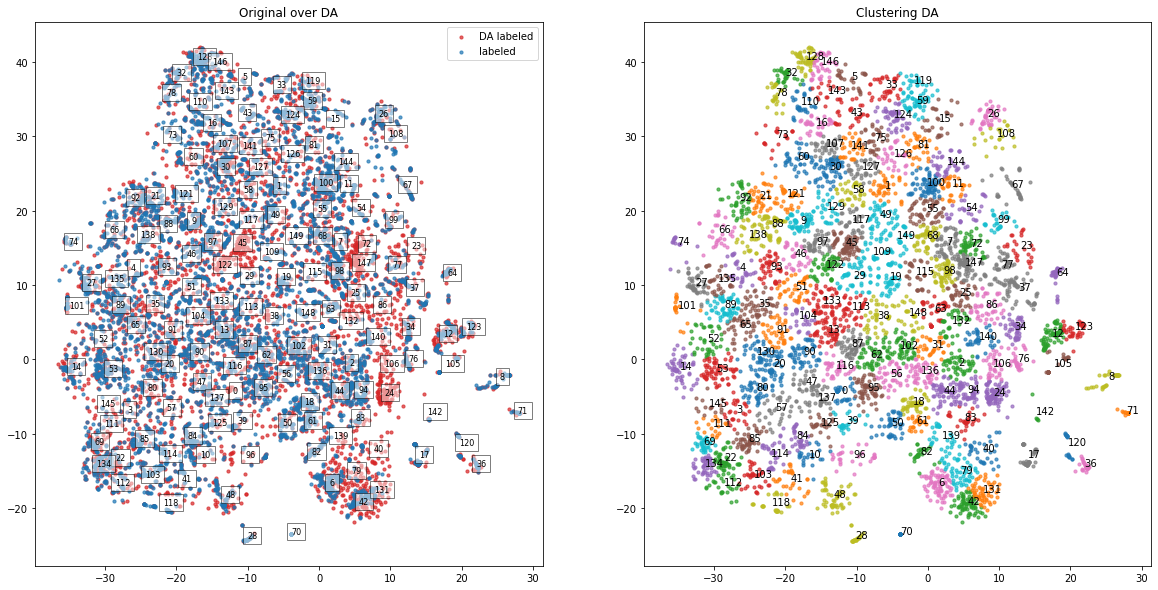}
    \caption{Left image is a t-SNE visualization of labeled and DA(labeled) at step 0. Right image is t-SNE visualization of DA(labeled) clustering region by K-means}
    \label{fig:tnse_labeled_da}

\end{figure*}

\begin{figure*}[ht]
    \centering
    \includegraphics[width=0.45\textwidth]{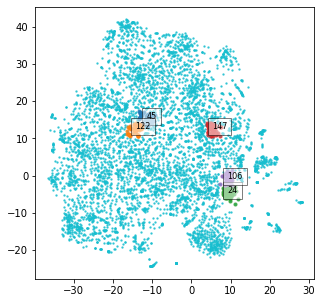}
    \includegraphics[width=0.6\textwidth]{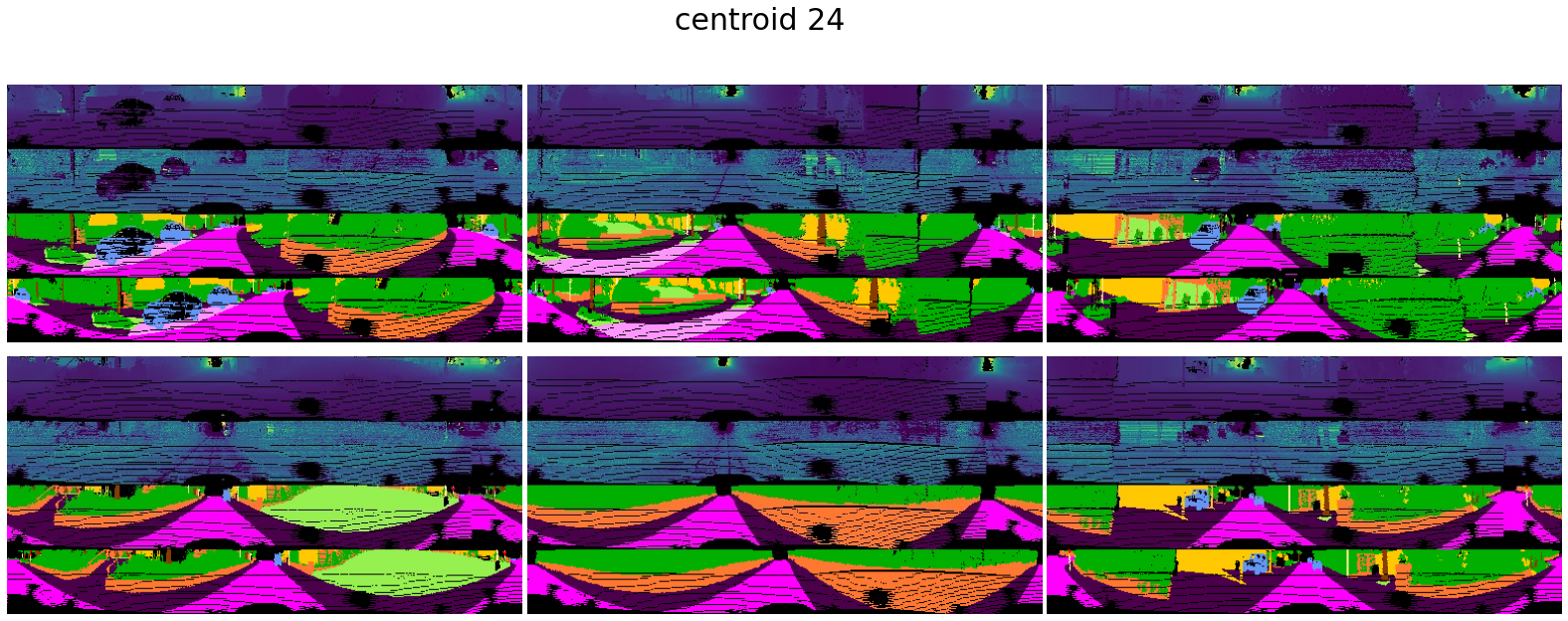}
    \includegraphics[width=0.6\textwidth]{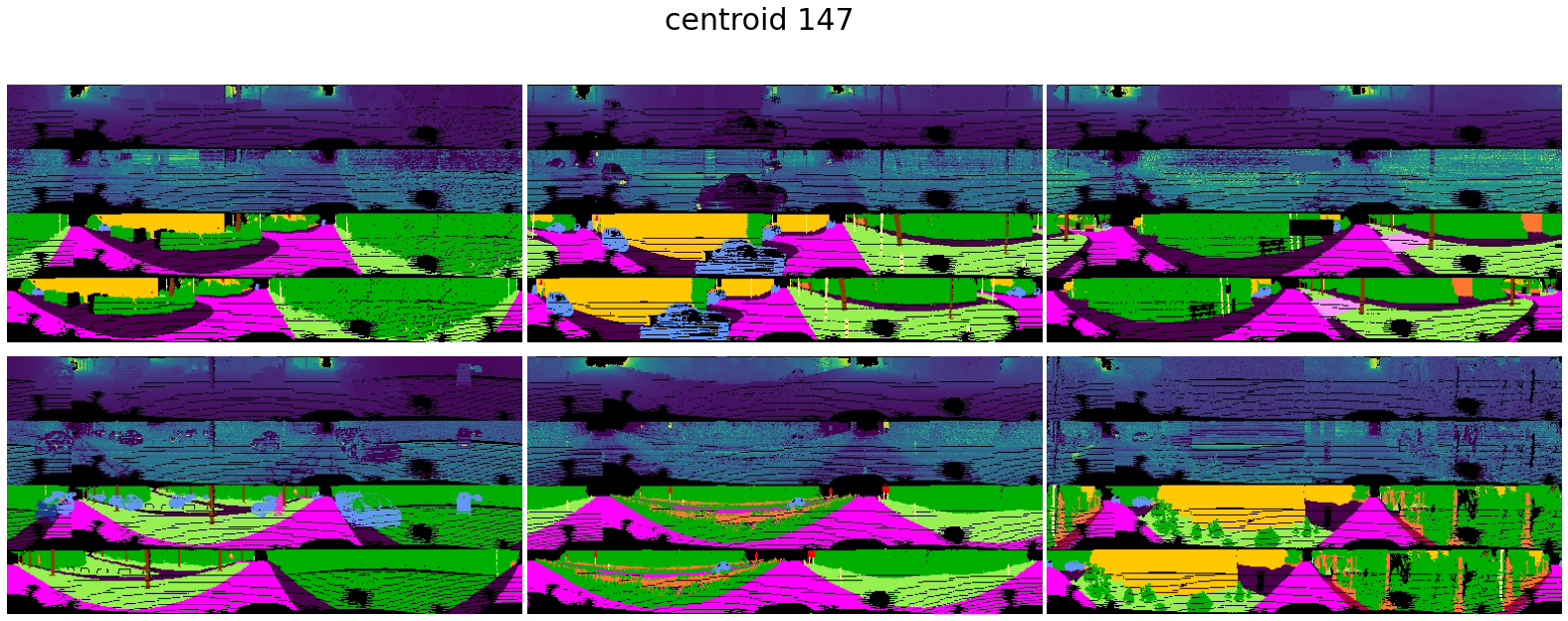}
    \includegraphics[width=0.6\textwidth]{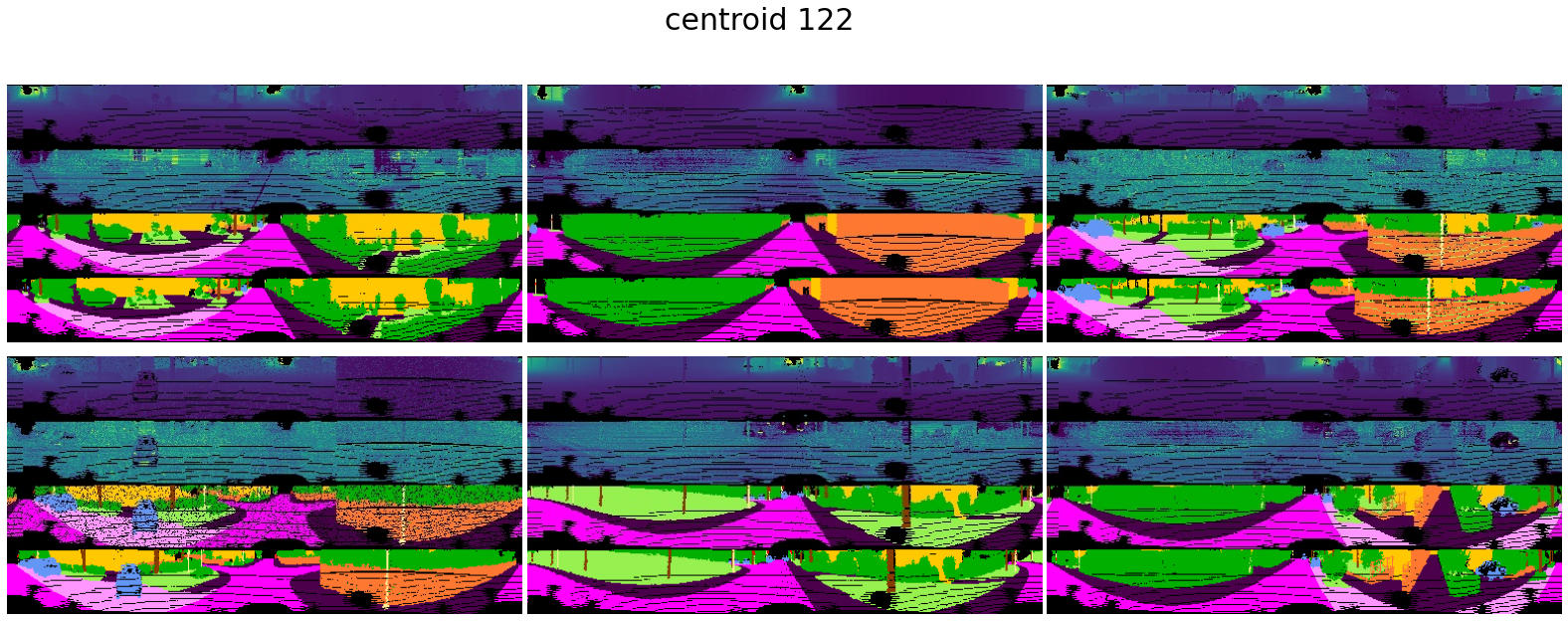}
    \includegraphics[width=0.6\textwidth]{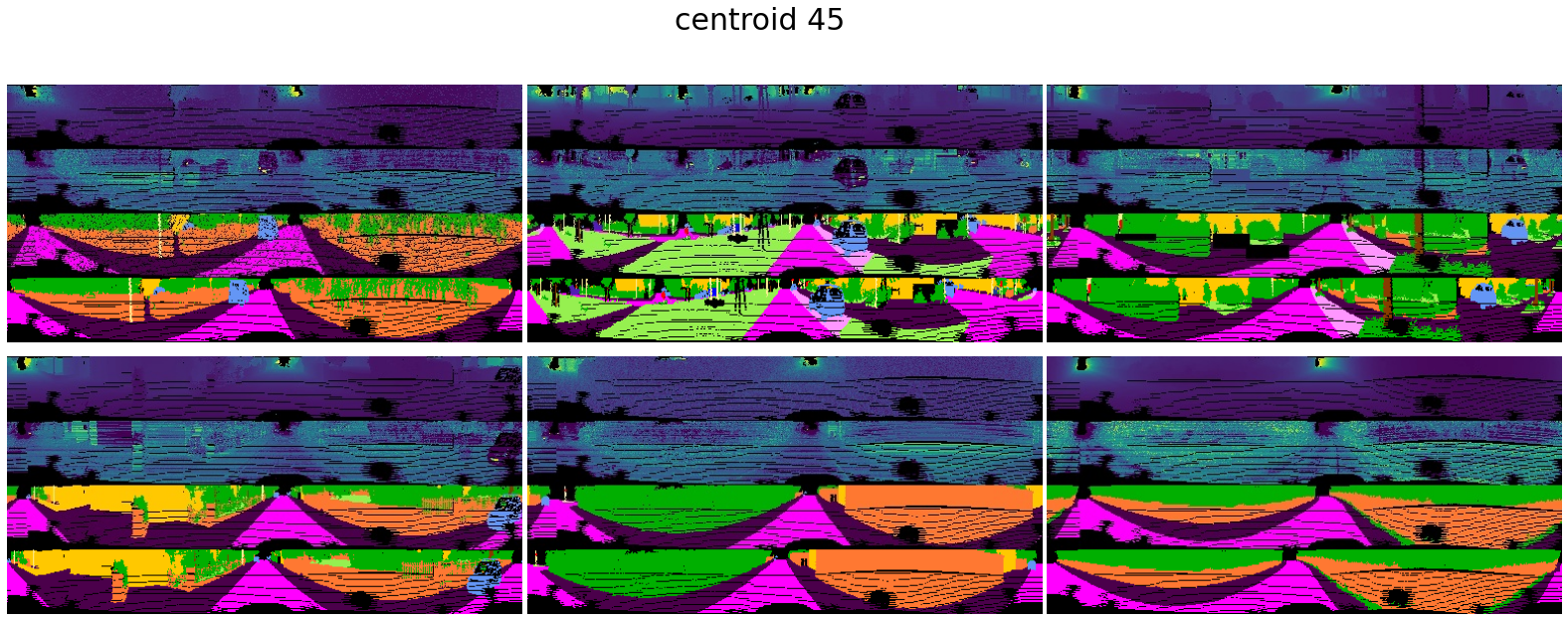}
    \caption{Visualization of samples deviating from dataset distribution.}
    \label{fig:tsne_labeled_da_centroids}
\end{figure*}

To visualize whether DA is relevant for Semantic-KITTI, we use t-SNE to reduce score images of labeled and DA (labeled) at AL step 0 of Semantic-KITTI/4's to 2D vector and plot on 2D axes (fig \ref{fig:tnse_labeled_da}). The samples located in regions with dense red color are out-of-training-set-distribution candidates at AL step 0. Visualization of the candidates is in fig \ref{fig:tsne_labeled_da_centroids}.

\begin{figure*}[ht]
    \centering
    \includegraphics[width=0.8\textwidth]{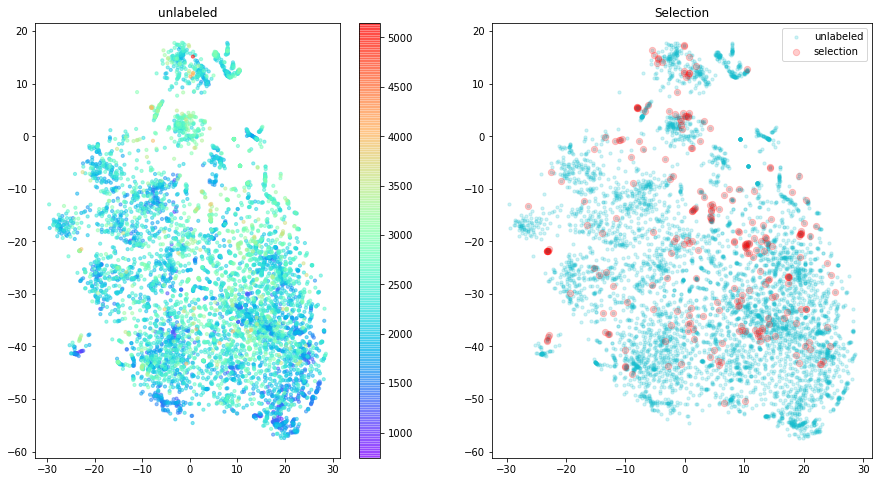}
    \caption{Left image is a t-SNE visualization of unlabeled pool at step 0, whose colors are corresponding to heuristic scores. Right image is the t-SNE visualization of selected samples and unlabeled pool. }
    \label{fig:tsne_selection}
\end{figure*}

In the figure \ref{fig:tsne_selection}, deep red spots show that there are a lot of selected high-score samples concentrated in the same location.

\section{Conclusion}
Core subset extraction/dataset distillation experiments were carried out on Semantic-KITTI dataset using SalsaNext and SSV2 models. The results of data augmentation and BALD heuristic were limited and did not perform any better than a random sampling method, the first baseline in active learning methods. These limited improvements are an important observation for active learning benchmarks, even though they constitute in marginal gains, and aids further studies to focus on other ways to improve labeling efficiency. Active learning strategies struggle with large-scale datasets such as point clouds in the real-world scenario.
\begin{enumerate}
    \item Data augmentation's have a strong effect on the quality of the heuristic function. It is thus key to evaluate data augmentation schemes that are dataset dependant, especially when unlabeled pool is just a small subset of target domain. Inappropriate or overwhelming augmentation can cause out-of-distribution, reducing model bias to target domain. \cite{7533048} propose an adaptive data augmentation algorithm for image classification, which looks for a small transform maximizing the loss on that transformed sample. 
    \item The final aggregation score in an active learning framework is a crucial choice. Aggregation is required when working on detection or segmentation tasks, while most AL study has been focused on classification. Here are a few choices of aggregation functions and how they affect the final ranking of images from which to sample:
        \begin{itemize}
            \item Sum of all scores: select a sample having a balance of high score and high number of elements.
            \item Average of all scores: select a sample having a high score comparable to the others regardless of the number of elements within a sample. 
            \item Max of all scores: select a sample having the highest element scores of all samples, targeting noises. 
            \item Weighted average of all scores: similar to average but focus on the given weights. The class weights can be  inverse-frequency class weights or customized based on the use case to prioritize the selection for certain classes.
        \end{itemize}
    It is noted that for our proposed method, sum and average eventually yield the same rankings because the number of pixels for range images are constant.
    \item Another key issue in industrial datasets is the filtering or exclusion of corrupted or outlier images/point clouds from the AL loop. \cite{chitta2019training} show that removing top highest uncertain samples as outliers improves model performances.
    \item To avoid selection of similar high-score samples and to reduce the sensitivity and bias of the model when training on a small dataset, a potential  strategy can be an integration of diversity. 
    Some hybrid methods, such as \cite{sener2018active} \cite{kirsch2019batchbald} \cite{ash2020deep} \cite{https://doi.org/10.48550/arxiv.2203.05053}, are shown to have more competitive results comparing to standalone uncertainty-based methods for datasets containing high rate of redundancy, but their computations are often costly for large scale point clouds datasets. Another strategy is using temporal information to sample \cite{https://doi.org/10.48550/arxiv.1911.09168} before or after uncertainty computation. Although it has low computation cost, it cannot ensure the dissimilarity among different samples.
    \item According to authors \cite{chitta2019training} \cite{https://doi.org/10.48550/arxiv.1811.03897}, MC Dropout is observed to lack of diversity for uncertainty estimation. To address this problem, \cite{chitta2019training} \cite{https://doi.org/10.48550/arxiv.1908.11757} use multiple checkpoints during training epochs from different random seeds.

    \item Pool-based AL might not be a proper AL strategy in real-world scenario for large scale datasets because of unlimited instances streamed sequentially, but limited at any given time. Due to memory limitation, it is impossible to store or re-process all past instances. In this case, stream-based methods are more pertinent by querying annotator for each sample coming from the stream. Although this type of query is often computationally inexpensive, the selected samples are not as informative as the ones selected by pool-based AL due to the lack of exploitation underlying distribution of datasets.
\end{enumerate}

\newpage

\textbf{Acknowledgements}
This work was granted access to HPC resources of [TGCC/CINES/IDRIS] under the
allocation 2021- [AD011012836] made by GENCI (Grand Equipement National de
Calcul Intensif). It is also part of the Deep Learning Segmentation (DLS)
project financed by \href{https://www.ademe.fr/}{ADEME}.

\ifCLASSOPTIONcaptionsoff
  \newpage
\fi

\bibliographystyle{apalike}
\bibliography{bibliography}

%








\end{document}